\documentclass[american,english,twoside]{IEEEtran}
\usepackage[T1]{fontenc}
\usepackage[latin9]{inputenc}
\usepackage{babel}
\usepackage{array}
\usepackage{rotating}
\usepackage{booktabs}
\usepackage{multirow}
\usepackage{graphicx}
\usepackage[unicode=true]
 {hyperref}

\makeatletter

\providecommand{\tabularnewline}{\\}

 \let\oldforeign@language\foreign@language
 \DeclareRobustCommand{\foreign@language}[1]{%
   \lowercase{\oldforeign@language{#1}}}

\usepackage{amsmath}
\usepackage{colortbl}

\DeclareMathOperator*{\argmax}{arg\,max}

\@ifundefined{showcaptionsetup}{}{%
 \PassOptionsToPackage{caption=false}{subfig}}
\usepackage{subfig}
\makeatother

\begin{document}

\title{Learning to Play Othello with Deep Neural Networks}

\author{Pawe\l{} Liskowski,~Wojciech~Ja\'{s}kowski,~Krzysztof Krawiec\thanks{The authors are with the Institute of Computing Science, Poznan University
of Technology, Poznan, Poland. W. Ja\'{s}kowski did this work while
staying at IDSIA Dalle Molle Institute for Artificial Intelligence
Research, Manno TI, Switzerland; e-mails: \{\protect\href{mailto:pliskowski@cs.put.poznan.pl}{pliskowski},
\protect\href{mailto:wjaskowski@cs.put.poznan.pl}{wjaskowski}, \protect\href{mailto:kkrawiec@cs.put.poznan.pl}{kkrawiec}\}@cs.put.poznan.pl.}}

\markboth{IEEE Transactions on Computational Intelligence and AI in Games}{Pawe\l ~Liskowski \MakeLowercase{\emph{et al.}}: Learning to Play
Othello with Deep Neural Networks}
\maketitle
\begin{abstract}
Achieving superhuman playing level by AlphaGo corroborated the capabilities
of convolutional neural architectures (CNNs) for capturing complex
spatial patterns. This result was to a great extent due to several
analogies between Go board states and 2D images CNNs have been designed
for, in particular translational invariance and a relatively large
board. In this paper, we verify whether CNN-based move predictors
prove effective for Othello, a game with significantly different characteristics,
including a much smaller board size and complete lack of translational
invariance. We compare several CNN architectures and board encodings,
augment them with state-of-the-art extensions, train on an extensive
database of experts' moves, and examine them with respect to move
prediction accuracy and playing strength. The empirical evaluation
confirms high capabilities of neural move predictors and suggests
a strong correlation between prediction accuracy and playing strength.
The best CNNs not only surpass all other $1$-ply Othello players
proposed to date but defeat ($2$-ply) Edax, the best open-source
Othello player.
\end{abstract}


\section{Introduction}

Neural networks, particularly the deep and convolutional ones (CNNs),
excel at recognizing virtually all kinds of patterns \cite{schmidhuber2015deep}.
They naturally generalize to arbitrary dimensionality, which makes
them suitable for analyzing raster images, time series, video sequences,
and 3D volumetric data in medical imaging \cite{szkulmowski2016oct}.
These properties make CNNs powerful for solving a wide range of challenging
tasks at unprecedented performance level, ranging from image classification
\cite{he2016deep}, to object localization \cite{girshick2014rich},
object detection \cite{redmon2016you}, image segmentation \cite{liskowski2016segmenting},
and even visual reinforcement learning in 3D environments \cite{Kempka2016ViZDoom}.

The structural analogy between 2D raster images and board states in
games is natural, so no wonder that it has been exploited within
the neural paradigm in the past (e.g., \cite{Dries2012Neural}) .
In most cases however, neural nets served there as \emph{board evaluation
functions} employed for searching game trees. It is only recently
that massive computing and efficient learning algorithms for deep
CNNs enabled their use as direct \emph{move predictors} capable of
achieving the level of play in the most challenging games previously
thought to be exclusive only to human players. This gave rise to a
line of Go-playing programs, including the paramount achievement of
AlphaGo, the superhuman-level Go-playing program \cite{silver2016mastering}. 

CNNs' success in Go was possible because it has been known for a long
time that Go players, rather than performing a ``mental'' tree search
of future game states, rely heavily on detecting patterns in board
states. This, together with a high branching factor of 250, and a
large board renders tree search approaches ineffective for Go. 

Here we ask whether CNNs have a practical potential for games of a
small branching factor and much smaller board size, for which the
minimax-style tree search algorithms perform well and achieve human-level
performance. To this aim, we consider learning CNN-based move predictors
for the game of Othello, a long-standing benchmark for AI \cite{Jaskowski2016cocmaes}.
This problem not only diverges from Go, but is also very different
from analysis of images in computer vision: the input rasters are
tiny ($8\times8$ ternary board states), free from noise and distortions
typical for real-world settings (sensor noise, lighting, perspective,
etc.), and the desirable invariances involve axial symmetries rather
than translation and scaling. Last but not least, every single piece
on the board matters and influences the decision. CNNs have been hardly
ever evaluated in such settings.

Our contributions include i) an experimental study of different CNN-based
architectures of deep neural networks for Othello; ii) state-of-the-art
move prediction accuracy on the French Othello league game dataset
WThor; iii) state-of-the-art $0$-ply policy for Othello (no look
ahead) that significantly improves over the previous approaches; iv)
an in-depth analysis of the characteristics of trained policies, including
the relationship between move prediction accuracy and winning odds,
strength and weaknesses at particular game stages, and confrontation
with opponents that employ game tree search at different depths.
The gathered evidence suggests that several of the proposed neural
architectures are best-to-date move predictors for Othello.

\section{Related Work\label{sec:Related-Work}}

CNNs have been introduced by Fukushima in Neocognitron \cite{fukushima1982neocognitron}
and occasionally used for selected image analysis tasks for the next
two decades \cite{lecun1998gradient}. The attempts of applying them
to broader classes of images remained however largely unsuccessful.
The more recent advent of deep learning brought substantial conceptual
breakthroughs, enabling training of deeper and more complex networks.
This, combined with cheap computing power offered by GPUs revolutionized
machine perception and allowed achieving unprecedented classification
accuracy, exemplified by the chain of ever-improving neural architectures
assessed on the ImageNet database \cite{krizhevsky2012imagenet,he2016deep}.

As signaled in the Introduction, the structural analogy between
natural images and game boards is quite obvious for humans and as
such could not remain unnoticed for long. This led to early attempts
of applying CNNs to board games \cite{clark2014teaching}. Soon it
became clear that the deep learning paradigm may be powerful enough
to be used together with game tree search techniques. This claim led
to DeepMind's AlphaGo, the first superhuman-level Go playing system
\cite{silver2016mastering}, which combined supervised learning, policy
gradient search and Monte Carlo Tree Search.

 Othello has for a long time been a popular benchmark for computational
intelligence methods \cite{Rosenbloom1982world,smith93coadaptive,Buro1997EvalOthello,lucas06temporal,Manning2010NashOthello,Dries2012Neural,Szubert2013scalability,Ree2013,Jaskowski2014systematic,Runarsson2014Preference,Jaskowski2016cocmaes}.
All strong Othello-playing programs use a variant of the minimax search
\cite{Buro1997OthelloMultiProbCut} with a board evaluation function.
Past research suggested that more can be gained by improving the latter
than the former; that is why recently the focus was mostly on training
$1$ look-ahead (a.k.a. $1$-ply) agents using either self-play \cite{Manning2010NashOthello,Jaskowski2016cocmaes},
fixed opponents \cite{Krawiec2011Learning,Jaskowski2014systematic},
or expert game databases \cite{Runarsson2014Preference}. Multiple
ways of training the agents have been proposed: value-based temporal
difference learning \cite{lucas06temporal,Dries2012Neural,Skoulakis2012},
(co)evolution \cite{Manning2010NashOthello,Samothrakis2012Coevolving,Jaskowski2013improving,Jaskowski2014ICGAsystematic},
and hybrids thereof \cite{szubert11learning,Szubert2013scalability}.

Designing a board evaluation function involves finding a good function
approximator. One of the most successful function approximators for
board games are \emph{$n$-tuple networks}. Interestingly, they were
originally proposed  for optical character recognition \cite{Bledsoe1959Pattern},
and their reuse for board games is yet another sign that the analogy
between images and piece arrangements on a board is feasible. They
were employed for the first time for Othello under the name of \emph{tabular
value functions} by Buro \cite{Buro1997OthelloMultiProbCut} in his
famous Logistello program and later popularized by Lucas \cite{lucas2007learning}.
Neural networks have also been used for Othello as function approximators,
e.g., in \cite{Dries2012Neural}, but, to the best of our knowledge,
CNNs have never been used for this purpose.

As we are not aware of any published move predictors for Othello,
in the experimental part of this paper we compare our method to a
representative sample of $1$-ply strategies. A comparison of the
state-of-the-art 1-ply players for Othello can be found in \cite{Jaskowski2016cocmaes}.
Among the published works, the best 1-ply player to date was obtained
by Coevolutionary CMA-ES (covariance matrix adaptation evolution strategies)
\cite{Jaskowski2016cocmaes} and systematic n-tuple networks. Among
the multi-ply Othello strategies, one of the contemporary leaders
is Edax\footnote{http://abulmo.perso.neuf.fr/edax/}, an open source
player that uses tabular value functions, one for each stage of the
game. Edax is highly optimized (bitboard move generator) and uses
deep tree minimax-like search (negasout with multi-probcut tree search
\cite{buro1997experiments}). Unfortunately, a description of how
its evaluation function was obtained is not publicly available.

One of the other top-5 Othello players to date is RL14\_iPrefN that
has been trained using preference learning on the French Othello League
expert games dataset \cite{Runarsson2014Preference}, reportedly obtaining
$53.0\%$ classification accuracy. In this paper, we significantly
improve over this baseline. 

\section{Othello}

\selectlanguage{american}%
\begin{figure}
\selectlanguage{english}%
\begin{centering}
\subfloat[\foreignlanguage{american}{\label{fig:Othello-initial-state}\foreignlanguage{english}{Othello
initial board state. Black to move.}}]{\begin{centering}
\includegraphics[width=0.2\textwidth]{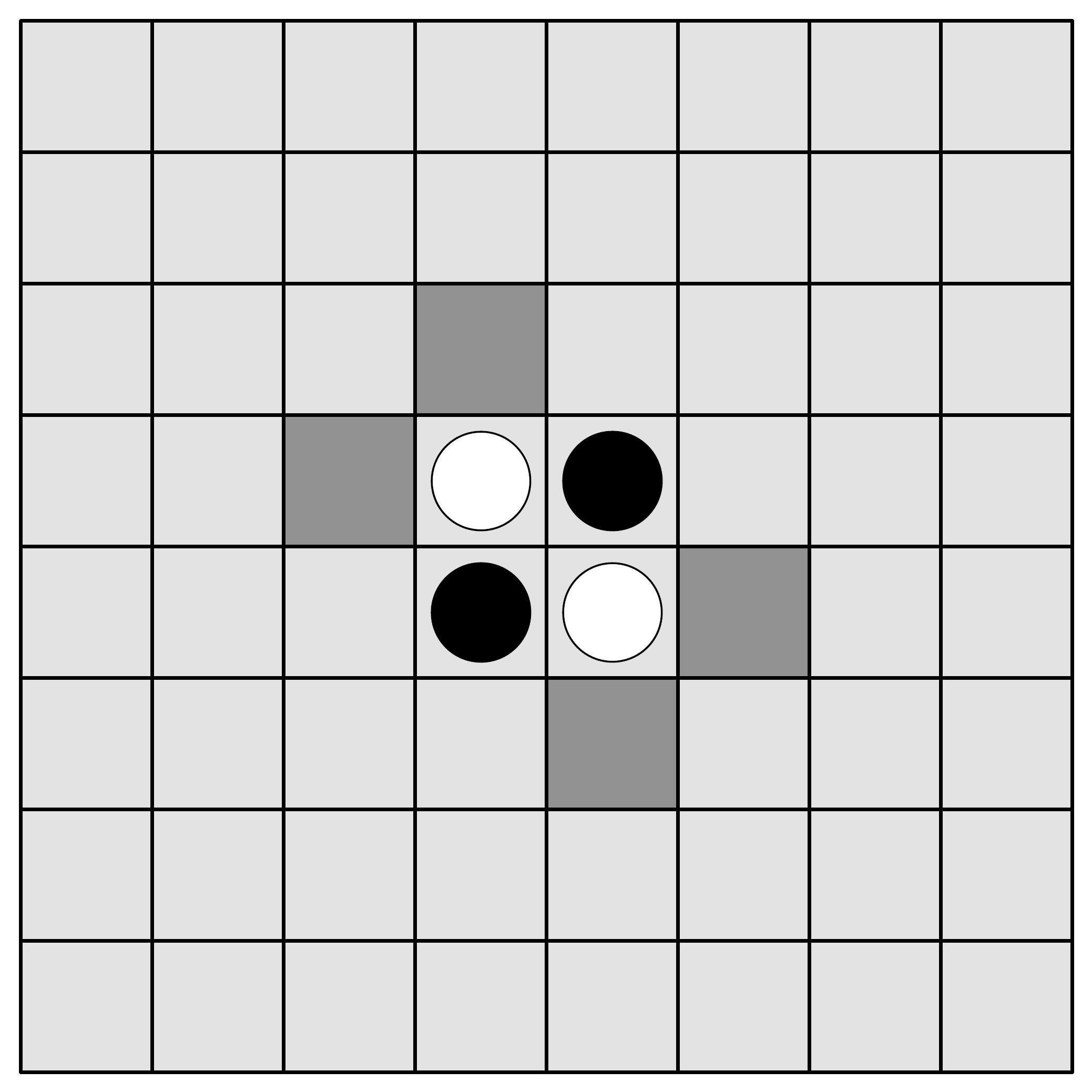}
\par\end{centering}
}\hfill{}\subfloat[Board state after black's move. White to move.\label{fig:Board-state-after-first-move}]{\begin{centering}
\includegraphics[width=0.2\textwidth]{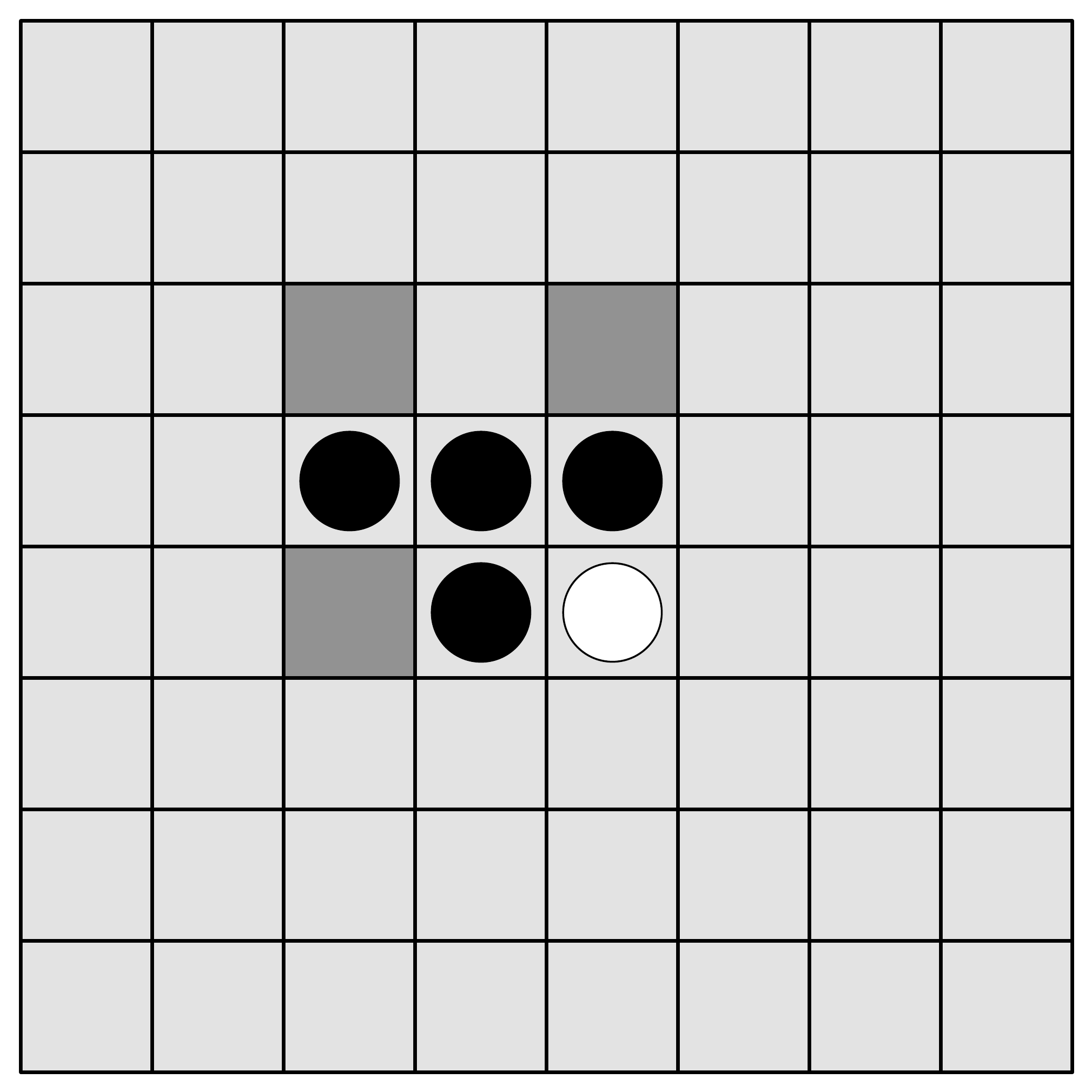}
\par\end{centering}
}
\par\end{centering}
\caption{Othello boards with legal moves marked as shaded locations.}
\selectlanguage{american}%
\end{figure}
\foreignlanguage{english}{Othello is a perfect information, zero-sum,
two-player strategy game played on an $8\times8$ board. There are
64 identical pieces which are white on one side and black on the other.
The game begins with each player having two pieces placed diagonally
in the center of the board (Fig. \ref{fig:Othello-initial-state}).
The black player moves first, placing a piece on one of four shaded
locations, which may lead to the board state in Fig. \ref{fig:Board-state-after-first-move}.
A move is legal if the newly placed piece is adjacent to an opponent\textquoteright s
piece and causes one or more of the opponent's pieces to become enclosed
from both sides on a horizontal, vertical or diagonal line. The enclosed
pieces are then flipped. Players alternate placing pieces on the board.
The game ends when neither player has a legal move, and the player
with more pieces on the board wins. If both players have the same
number of pieces, the game ends in a draw. }

\selectlanguage{english}%
Despite simple rules, the game of Othello is far from trivial. The
number of legal positions is approximately $10^{28}$ and the game
tree has in the order of $10^{58}$ nodes \cite{allis1994searching},
which precludes any exhaustive search method. Othello is also characterized
by a high temporal volatility: a high number of pieces can be flipped
in a single move, dramatically changing the board state. These features
and the fact that the game has not yet been solved makes Othello an
interesting test-bed for computational intelligence methods.

\section{Learning to Predict Expert Moves \label{sec:Method}}

\subsection{The Classification Problem }

\begin{figure*}[t]
\centering{}\includegraphics[scale=0.78]{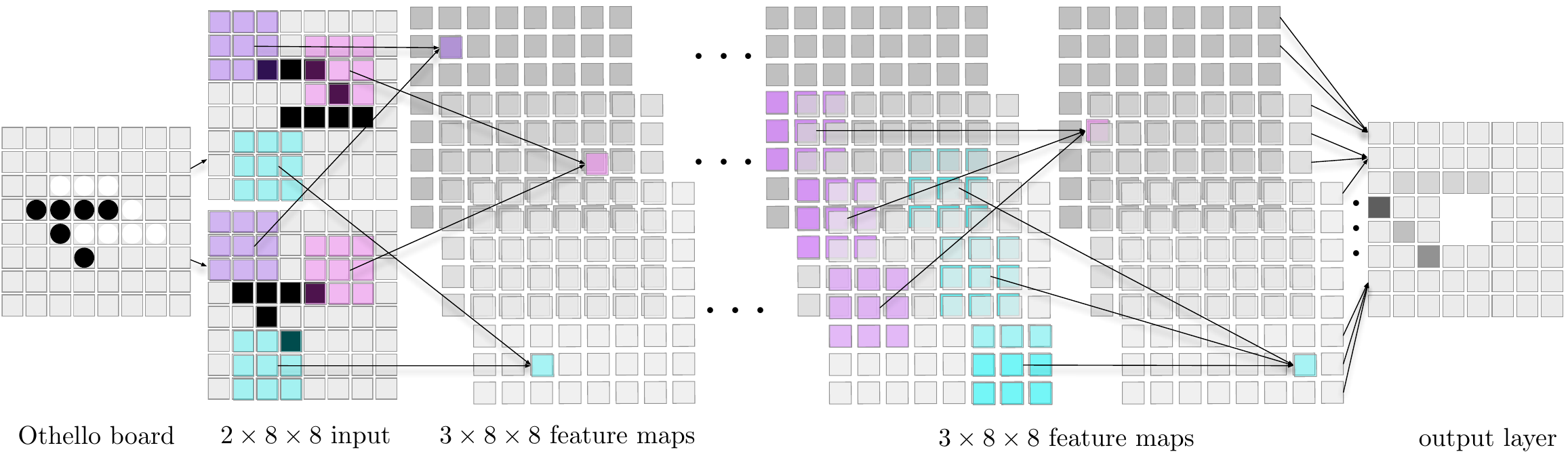}\caption{\label{fig:cnn-architecture}Exemplary CNN for move prediction. The
board is fed into the network as an $8\times8$ two-channel binary
image, where ones indicate the locations of the player's (first channel)
and opponent's pieces (second channel). Units in a feature map (colored
squares) receive data from $3\times3$ RFs (in the same color) in
the image (or in a previous layer). The network uses zero-padding
on the edges (not shown),  so maps in all layers are $8\times8$.
For clarity, each hidden layer has only $3$ feature maps.}
\vspace{-5mm}
\end{figure*}
Since Othello is a deterministic and fully-observable environment,
a policy $\pi:\ X\rightarrow Y$ is here a mapping from the space
of board states $X$ to the space of actions $Y=\{1,\dots,K\}$ that
correspond one-to-one to the $K=60$ board positions. $\pi(x)$ represents
the desired move for board state $x\in X$. Given a training set of
$N$ examples of good (e.g., experts') moves in $X\times Y$, we can
formulate the problem of finding $\pi$ as a \emph{multi-class classification
}problem. A typical approach to such problems is training a parameterized
probability function $p(y|x;\theta)$, that represents the confidence
of making a move $y$ in state $x$. To estimate $p$, one needs to
minimize the cross-entropy loss:
\[
L(\theta)=-\sum_{i=1}^{N}\sum_{k=1}^{K}y_{i}^{(k)}\ln p(y_{i}^{(k)}|x_{i};\theta),
\]
where $y^{(k)}=1$ if $y=k$ and $0$ otherwise. The task of a training
algorithm is to estimate the optimal $\theta$ and so find $\hat{\theta}$,
a policy parameterization that brings $L(\theta)$ sufficiently close
to its global optimum. Once found, the player chooses the moves of
highest confidence:
\[
\pi_{\hat{\theta}}(x)=\argmax_{y\in Legal(x)}p(y|x;\hat{\theta}),
\]
where $Legal(x)\subset Y$ is the subset of moves that are legal
in board state $x$. 

If $\pi$ determines the move to make without explicit simulation
of the future game course (using, e.g., minimax-style algorithms),
we refer to it as \emph{move predictor}. One-to-one correspondence
between board locations and moves in Othello greatly facilitates the
design of move predictors. For other board games that involve heterogeneous
pieces that move around the board, the space of moves $Y$ is more
complex and requires a more sophisticated representation.

\subsection{CNNs for Move Prediction}

\begin{table*}[t]
\caption{\label{tab:architecture}CNN architectures used in the experiments.
Layer names are followed by numbers of feature maps. }

\centering{}\renewcommand{\arraystretch}{0.7}\foreignlanguage{american}{\footnotesize

\begin{tabular}{p{8mm}p{16cm}}
\toprule
Conv4 &  ${conv64}\rightarrow {conv64}\rightarrow {conv128}\rightarrow {conv128}\rightarrow fc128\rightarrow fc60$ \\
\\
Conv6 &  ${conv64}\rightarrow {conv64}\rightarrow {conv128}\rightarrow {conv128}\rightarrow {conv256}\rightarrow {conv256}\rightarrow fc128\rightarrow fc60$ \\
\\
Conv8 &  ${conv64}\rightarrow {conv64}\rightarrow {conv128}\rightarrow {conv128}\rightarrow {conv256}\rightarrow {conv256}\rightarrow {conv256}\rightarrow {conv256}\rightarrow fc128\rightarrow fc60$ \\

\bottomrule
\end{tabular}

}
\end{table*}
In the simplest scenario, the board states in $X$ could be encoded
as $64$ ternary variables and fed into any probability estimator.
However, generic probability estimators are not biased towards the
kind of structure that relates board state variables in Othello. To
see that such a structure exists, it is enough to imagine the board
variables being randomly permuted, which would completely obfuscate
the game for human players (while being still the same policy learning
task under the formulation in the previous section). This shows that
capturing spatial patterns of the player's and opponent's pieces is
essential, and makes it particularly appropriate to take advantage
of learning systems that have such capabilities, i.e., treat a board
state as an image. 

Convolutional neural networks (CNNs) are nowadays the leading machine
learning tool for image analysis (cf. the work cited in the introduction).
In the following, we briefly describe the CNN architecture as we tailored
it to move prediction in Othello, referring readers interested in
this topic to other comprehensive studies \cite{bengio2009learning,schmidhuber2015deep}.
For clarity, we present here only the core CNN architecture and augment
it with extensions in the subsequent experimental section. 

A CNN is a composite of multiple elementary processing units, each
equipped with weighted inputs and one output, performing convolution
of input signals with weights and transforming the outcome with some
form of nonlinearity. The units are arranged in rectangular \emph{feature
maps} (grids) and spatially aligned with the raster of the input image.
The spatial arrangement of units is the primary characteristic that
makes CNNs suitable for processing visual information. Feature maps
are arranged into \emph{layers}, where maps in a given layer fetch
signals only from the previous layer (see also Fig. \ref{fig:cnn-architecture}).

Our CNNs take as input the representation of the current board state,
encoded in two $8\times8$ binary matrices. The $1$s indicate the
locations of the player's pieces in the first matrix and opponent's
pieces in the second matrix.  We refer to this encoding as \emph{pieces}.
In the encoding called \emph{vmoves}, we add a third matrix in which
we mark with $1$s the player's legal moves, to facilitate picking
the valid moves during game playing. Following recent research conducted
in Go \cite{silver2016mastering}, we define also the $ones$ encoding,
which includes an additional regularizing channel in \emph{pieces}
encoding that contains only ones.

A board state represented in the above way forms an $8\times8$ binary
image comprising two (for \emph{pieces}), or three (for \emph{vmoves}
and \emph{ones}) channels, and is fetched by the feature maps in the
first layer of the CNN. Each unit in a feature map receives data only
from its \emph{receptive field} (RF), a small rectangle of neighboring
board locations, however from all channels, and uses a separate weight
set for each channel (see exemplary RFs in Fig. \ref{fig:cnn-architecture}).
Analogously in the subsequent layers, a unit fetches data from the
corresponding RFs in all feature maps in the previous layer, and treats
them with separate weight sets. 

The RFs of neighboring units in a map are offset by a \emph{stride}.
Board size, RF size and stride together determine the dimensions of
a feature map. For instance, a feature map with $3\times3$ RFs with
stride $1$ needs only $36$ units when applied to an $8\times8$
board, because six RFs of width $3$ each, overlapping by two rows,
span the entire board width (and analogously for height). The local
connectivity substantially reduces the number of weights and facilitates
capturing of spatial patterns. 

Units within the same feature map share their weights, and so calculate
the same local feature, albeit from a different part of the board.
This reduces the number of parameters even further and makes the extracted
features equivariant. For instance, the $6\times6$ feature map mentioned
above has only $9\cdot b+1$ parameters for a $b$-channel input ($3\times3=9$
plus one threshold per feature map). 

We use rectified linear units (ReLU) to process convolution outcomes,
with the nonlinearity defined as $f(x)=max(x,0)$. ReLUs make networks
learn faster than squeezing-function units (e.g., sigmoid or $\tanh$)
and are particularly effective in networks with many layers, effectively
alleviating the vanishing gradient problem, because gradient backpropagates
through them undistorted for all positive excitations, in contrast
to a small interval of excitations for the squeezing units.

Our basic architectures are the networks composed of $4$, $6$, and
$8$ convolutional layers shown in Table \ref{tab:architecture}.
In each architecture, the input is passed through a stack of convolutional
layers composed of 64, 128, or 256 feature maps, with $3\times3$
RFs and the stride set to $1$. Since the input is very small, we
pad it with zeros on the grid border (so that the RFs can reach also
beyond the input grid) in order to preserve the spatial resolution
after convolution. For the same reason, we do not use any form of
\emph{pooling}, i.e., aggregating the outputs of multiple units by
other means than convolution, as there is little rationale for that
for small input grids \cite{springenberg2014striving}. Consequently,
all feature maps in all layers have $8\times8$ units. 

In all architectures, the stack of convolutional layers is followed
by two fully-connected (FC) layers. The hidden layer consists of $128$
units with ReLU activations, while the output layer has $K=60$ units
corresponding to $60$ possible move candidates. Each output $y^{(k)}$
corresponds to a single board location, and the desired moves are
encoded using the binary $1$-of-$K$ scheme (a.k.a. \emph{one-hot
encoding}). For the output values to be interpreted as probabilities
they must sum to unity, which is achieved using the softmax transformation:
\[
o_{k}=\frac{\mathrm{e}^{y^{(k)}}}{\sum_{j}\mathrm{e}^{y^{(j)}}},
\]

\noindent The classifier selects the move with the highest probability,
i.e., $\hat{o}=\argmax_{k\in K}o_{k}$.

The architectures presented in Table \ref{tab:architecture} resulted
from a series of preliminary experiments, where we found out that
increasing network depth beyond eight convolutional layers does not
improve the performance while significantly lengthening the training
time. Also, small $3\times3$ RFs and deeper networks turn out to
be more efficient in terms of trading-off runtime and accuracy than
using fewer but larger RFs.

\section{Experiments in Move Prediction Accuracy\label{sec:ExpMovePred}}

\subsection{Experimental Setup\label{sec:Setup}}

In this section, we consider networks as move predictors and do not
engage them in actual games. The key performance indicator is prediction
accuracy, i.e. the percentage of correctly predicted moves. We compare
various configurations of the neural move predictors described in
Section \ref{sec:Method} trained on examples of moves extracted from
games played by humans (Section \ref{subsec:Training}) . 

\subsubsection{Datasets}

Created in 1985, the WThor database\footnote{http://www.ffothello.org/informatique/la-base-wthor/}
contains the records of $119{,}339$ games played by professional
Othello players in various French tournaments. From those records,
we extracted all board states accompanied with the color of the player
to move and the move chosen by the player in that particular state.
The resulting set of $6{,}874{,}503$ (\emph{board, color, move})
triples is referred to as \emph{Original}. We then removed the duplicates,
obtaining $4{,}880{,}413$ unique (\emph{board, color, move}) triples,
referred to as \emph{Unique} dataset in the following. Note that
\emph{both} datasets may contain inconsistent examples, i.e. the same
board state associated with different moves. A perfect classifier
can achieve accuracy of $91.16\%$ on \emph{Original} and $97.49\%$
on \emph{Unique}. The color attribute is used only to invert the pieces
on the boards when the white player is to move, so that all states
in the dataset are seen from the black player's perspective.

Deep neural networks are known to perform particularly well when trained
on large volumes of data. In order to augment the datasets, we take
advantage of Othello's invariance to board symmetries. For each board
state in the \emph{Original} and \emph{Unique} datasets, we produce
its all seven reflections and rotations. This results respectively
in two \emph{symmetric} \emph{datasets}: \emph{Original-S} ($54{,}996{,}024$
examples) and \emph{Unique-S} ($39{,}019{,}056$ examples), where
the latter was cleared also of the duplicates resulting from symmetric
transformations.

To assess the generalization capabilities of move predictors, we partition
each dataset into disjoint training and testing sets. For the\emph{
}asymmetric datasets (\emph{Original} and \emph{Unique}), we allocate
25\% of examples for testing. The symmetric datasets are much larger,
so we allocate only 5\% of examples for testing, so that all test
sets are similar in size.

\subsubsection{Training\label{subsec:Training}}

For training, we use stochastic gradient descent with $0.95$ momentum.
Units' weights are initialized using He's method \cite{he2015delving}
while the biases (offsets) are initialized with zeroes. The learning
rate is initially set to $0.1$, and then halved twice per epoch.
The loss function is regularized by the $L_{2}$ norm with the weight
of $5\cdot10^{-4}$. Learning is stopped after $24$ epochs of training
for the asymmetric datasets ($171,600$ batches of examples) and $6$
epochs ($434,400$ batches) for the symmetric datasets.

The implementation is based on the Theano framework \cite{theanoArxiv},
which performs all computation on a GPU in single-precision arithmetic.
The experiments were conducted on an Intel Core i7-4770K CPU with
NVIDIA GTX Titan GPU. Training time varied from single hours to dozens
of hours depending on configuration, e.g. nearly $62$ hours for Conv8
trained on \emph{Unique-S}.

\subsection{Results\label{subsec:MovePredResults}}

\subsubsection{Effect of Data Augmentation\label{subsec:PredictionDataAug}}

\begin{table}
\caption{\label{tab:augment}Prediction accuracy of networks trained on four
different training sets and evaluated on the \emph{Original} test
set. \emph{Pieces} encoding was used. }

\begin{centering}
\par\end{centering}
\centering{}%
\begin{tabular}{lcccc}
\hline 
\multirow{1}{*}{Architecture} & \emph{Original} & \emph{Original-S} & \emph{Unique} & \emph{Unique-S}\tabularnewline
\hline 
Conv4 & 55.1 & 56.9 & 55.9 & \textbf{57.2}\tabularnewline
Conv6 & 57.2 & 58.7 & 57.9 & \textbf{59.1}\tabularnewline
Conv8 & 58.1 & 60.1 & 58.9 & \textbf{60.5}\tabularnewline
\hline 
\end{tabular}
\end{table}

Table \ref{tab:augment} presents the test-set prediction accuracy
of three basic network architectures trained on particular training
sets with the \emph{pieces} encoding. For fairness, the networks are
compared on the same data, i.e. the \emph{Original }test set. Training
on the datasets augmented by symmetries systematically leads to a
higher prediction accuracy. The impact of removing duplicates is also
consistently positive, albeit not so strong. These two trends together
render \emph{Unique-S} most promising, so in the following experiments
we use this dataset only.

\subsubsection{Impact of Board Encoding and the Number of Layers}

\begin{table}
\caption{\label{tab:results-encoding}Prediction accuracy on \emph{Unique-S}
test set for different board encodings and number of convolutional
layers.}

\begin{centering}
\renewcommand{\arraystretch}{0.7}
\par\end{centering}
\centering{}%
\begin{tabular}{lccc}
\toprule 
Architecture & \emph{pieces} & \emph{ones} & \emph{vmoves}\tabularnewline
\midrule
Conv4 & 57.2 & \textbf{57.3} & \textbf{57.3}\tabularnewline
Conv6 & \textbf{59.2} & 59.1 & 59.0\tabularnewline
Conv8 & 60.5 & \textbf{60.6} & 59.0\tabularnewline
\bottomrule
\end{tabular}
\end{table}

Table \ref{tab:results-encoding} presents the prediction accuracy
of the three architectures when trained on \emph{Unique-}S with various
encodings of the board state. This time, the networks are tested on
the \emph{Unique-S} test set, which explains the minor differences
w.r.t. Table \ref{tab:augment}, where \emph{Original} test set was
used for testing. The accuracy clearly correlates with the number
of layers, so it is tempting to reach for even deeper architectures.
However, a $10$-layer architecture analogous to Conv8 failed to
further improve the prediction accuracy. 

Concerning the comparison of encodings, \emph{pieces} and \emph{ones}
seem to perform on par, so the former should be preferred as it does
not require an additional input layer/channel. The relative underperformance
of \emph{vmoves} is more puzzling as it is hard to see how giving
a learner a hint about moves' validity could deteriorate its performance.
In an attempt to investigate this outcome, we evaluate the networks
with respect to their ability to discern the valid and invalid moves.
In Table \ref{tab:results-valid-move-acc}, we present the test-set
accuracy of particular networks, where $100$\% indicates that the
network's most excited output is always among the valid moves. All
considered networks and encodings bring this indicator very close
to perfection, with the networks using \emph{vmoves} making no mistakes
at all in that respect. This suggests that telling apart the valid
moves from the invalid ones is very easy regardless of configuration,
and providing this information explicitly in \emph{vmoves} is not
only unnecessary but may distract the training process. Therefore,
we use the \emph{pieces} encoding in all experiments that follow.

\begin{table}
\caption{\label{tab:results-valid-move-acc}Probability {[}\%{]} of choosing
a valid move on \emph{Unique-S}.}

\begin{centering}
\renewcommand{\arraystretch}{0.7}
\par\end{centering}
\centering{}%
\begin{tabular}{lccc}
\toprule 
Architecture & \emph{pieces} & \emph{ones} & \emph{vmoves}\tabularnewline
\midrule
Conv4 & 99.95 & 99.95 & 100.0\tabularnewline
Conv6 & 99.99 & 99.99 & 100.0\tabularnewline
Conv8 & 99.99 & 99.99 & 100.0\tabularnewline
\bottomrule
\end{tabular}
\end{table}

\subsubsection{Impact of Regularization and Bagging}

\begin{table}
\caption{\label{tab:results-reg}The influence of dropout, batch normalization,
and bagging on prediction accuracy {[}\%{]}. \emph{Pieces} encoding
was used. }

\begin{centering}
\renewcommand{\arraystretch}{0.7}
\par\end{centering}
\centering{}%
\begin{tabular}{lcc}
\toprule 
Architecture & Prediction accuracy & Prediction time {[}$\times10^{-4}$s{]}\tabularnewline
\midrule 
Conv8 (baseline) & 60.5 & 1.6 \tabularnewline
Conv8+dropout & 58.8 & 1.6\tabularnewline
Conv8+BN & \textbf{62.7} & 1.9 \tabularnewline
Conv8+BN+bagging & \textbf{64.0} & 19.4\tabularnewline
\bottomrule
\end{tabular}
\end{table}

In Table \ref{tab:results-reg}, we report the impact of a few techniques
that tend to improve generalization performance of CNNs. 

\emph{Dropout} consists in disabling a number of (here: $50$ percent)
randomly selected units for the individual batches of training examples,
which forces a network to form multiple alternative pathways from
inputs to outputs. This usually makes networks more robust and reduces
overfitting, particularly in domains where input data are inherently
redundant, like natural images, typically composed of at least thousands
of pixels. The states of Othello board are however very different:
there are only $64$ inputs, and every single piece counts when choosing
the move to make. There is virtually no redundancy in the input. As
Table \ref{tab:results-reg} suggests, dropout prevents the Conv8
network from capturing the board state in full and so hampers its
predictive accuracy. 

The purpose of \emph{batch normalization }(BN) is to address the \emph{internal
covariate shift} \cite{ioffe2015batch} by standardizing the outputs
of individual network units over batches of examples, which was demonstrated
to be beneficial in many contexts. This is corroborated in our case
with over two percent gain in accuracy, so we consider this configuration
as a baseline throughout the rest of this paper. 

Finally, we attempt to reduce the classifier's variance by employing
bootstrap aggregation (\emph{bagging}). We train $10$ networks on
bootstrapped samples drawn from the training set and average their
corresponding outputs to determine the move to make. When applied
to the Conv8+BN network (Table \ref{tab:results-reg}), bagging boosts
the performance to $64\%$. The price paid for that is a 10-fold increase
in not only the training time\footnote{The mean prediction times were computed using batch size of 256. Making
decision for a \emph{single} board state takes on average 5-20 times
more GPU time.}, but also the prediction time, which will become relevant in the
later parts of the study, when confronting our strategies with other
opponents.

\subsubsection{Architectures with Skip-Connections}

The more recent DL research brought very deep network architectures
containing dozens of layers \cite{srivastava2015highway,he2016deep},
which became the state-of-the-art in image classification. Training
such networks is only possible due to\emph{ skip connections} that
form shortcuts in the pathways between layers, typically adding the
representation (feature map) learned in a given layer to the representation
learned in one of the consecutive layers. This, particularly when
repeated multiple times across the network, opens new ways to achieving
the training goal: rather than learning a mapping $H$ such that $y(x)=H(x)$,
a network learns a residual $F$ such that $y(x)=F(x)+x$. Skip connections
alleviates the vanishing gradient problem as the gradient can backpropagate
unchanged through many layers. 

Table \ref{tab:results-resnets} presents the predictive accuracy
of several skip-connection architectures. A ResNet-$n$ network \cite{he2016deep}
consists of $\frac{n}{2}$\emph{ residual blocks}, each comprising
a nonlinear layer (ReLU activation) followed by a linear layer, shortcut
with a skip connection that adds the input of the first layer to the
output of the second one. Originally designed for image analysis,
ResNets  perform downsampling that reduces the input dimensionality;
as there is arguably little need for that for our very small $8\times8$
input, we consider a variant stripped off that feature, ResNet-$n$-np.
As expected, maintaining the dimensionality of the input throughout
learning improves the accuracy of the smaller (ResNet-32) and larger
(ResNet-56) architecture. 

\begin{table}
\caption{\label{tab:results-resnets} Prediction accuracy of residual networks.}

\begin{centering}
\renewcommand{\arraystretch}{0.7}
\par\end{centering}
\centering{}%
\begin{tabular}{lcc}
\toprule 
Architecture & Accuracy &  Prediction time {[}$\times10^{-4}$s{]}\tabularnewline
\midrule
Conv8+BN (baseline) & \textbf{62.7} & 1.9\tabularnewline
ResNet-32 & 57.4 & 1.0\tabularnewline
ResNet-32-np & 59.2 & 1.0 \tabularnewline
ResNet-32-np-p & 59.5 & 1.0\tabularnewline
\midrule
ResNet-56 & 58.3 & 1.8 \tabularnewline
ResNet-56-np & 60.1 & 1.8 \tabularnewline
ResNet-56-np\textendash p & 60.5 & 1.8 \tabularnewline
\midrule
DenseNet-40 & 59.9 & 4.8 \tabularnewline
DenseNet-100 & \textbf{61.9} & 37.2 \tabularnewline
\midrule
WideNet-40-4 & \textbf{62.2} & 9.4 \tabularnewline
\bottomrule
\end{tabular}
\end{table}
Simple adding of layer's outputs is possible only when they have the
same dimensions. Otherwise, the authors of ResNets propose to implement
the skip connections as linear mappings. The ResNet-$n$-np-p configurations
in Table \ref{tab:results-resnets} implement that feature, which
slightly improves the predictions. 

ResNets' skip connections can be considered first-order as each of
them spans the inputs and outputs of the same residual block. In DenseNets
\cite{huang2016densely}, skip connections are being introduced between
each pair of layers, so in a network of $l$ layers (or residual blocks)
there are $\frac{l(l-1)}{2}$ of them. We trained two DenseNet architectures.
The smaller one (DenseNet-40) managed to perform on par with the other
networks considered in this section, while the larger one (DenseNet-100)
significantly outperformed them, almost reaching the performance level
of the baseline. 

Finally, a very recent work on architectures dubbed WideNet \cite{zagoruyko2016wide}
revealed that the number of layers (and thus parameters) in residual
networks can be significantly reduced with little or no harm for accuracy
by making the residual blocks more expressive. This can be achieved
by increasing the number of convolution layers within a block, increasing
the number of feature maps, or both. Our instance of this architecture
comprises $3$ stacks of residual blocks, each stack composed of
$6$ blocks. Each residual block contains two $3\times3$ convolutional
layers. The layers in consecutive stacks are equipped with respectively
$64$, $128$, and $256$ feature maps (cf. Table 1 in \cite{zagoruyko2016wide}
for $k=4$), so the network is four times wider than the original
ResNet. As the last row of Table \ref{tab:results-resnets} demonstrates,
this WideNet configuration fares the best among the skip-connections
architectures considered here, which suggests that aggregating multiple
features at various stages of processing is essential for successful
move prediction. However, WideNet-40-4 is also one of the slowest
networks when queried.

Overall however, contrary to the expectations, neither of the very-deep
skip-connection networks did improve over the Conv8+BN baseline.

\subsubsection{Prediction Accuracy Analysis}

\begin{figure}
\centering{}\includegraphics[bb=5bp 5bp 620bp 732bp,width=0.6\textwidth]{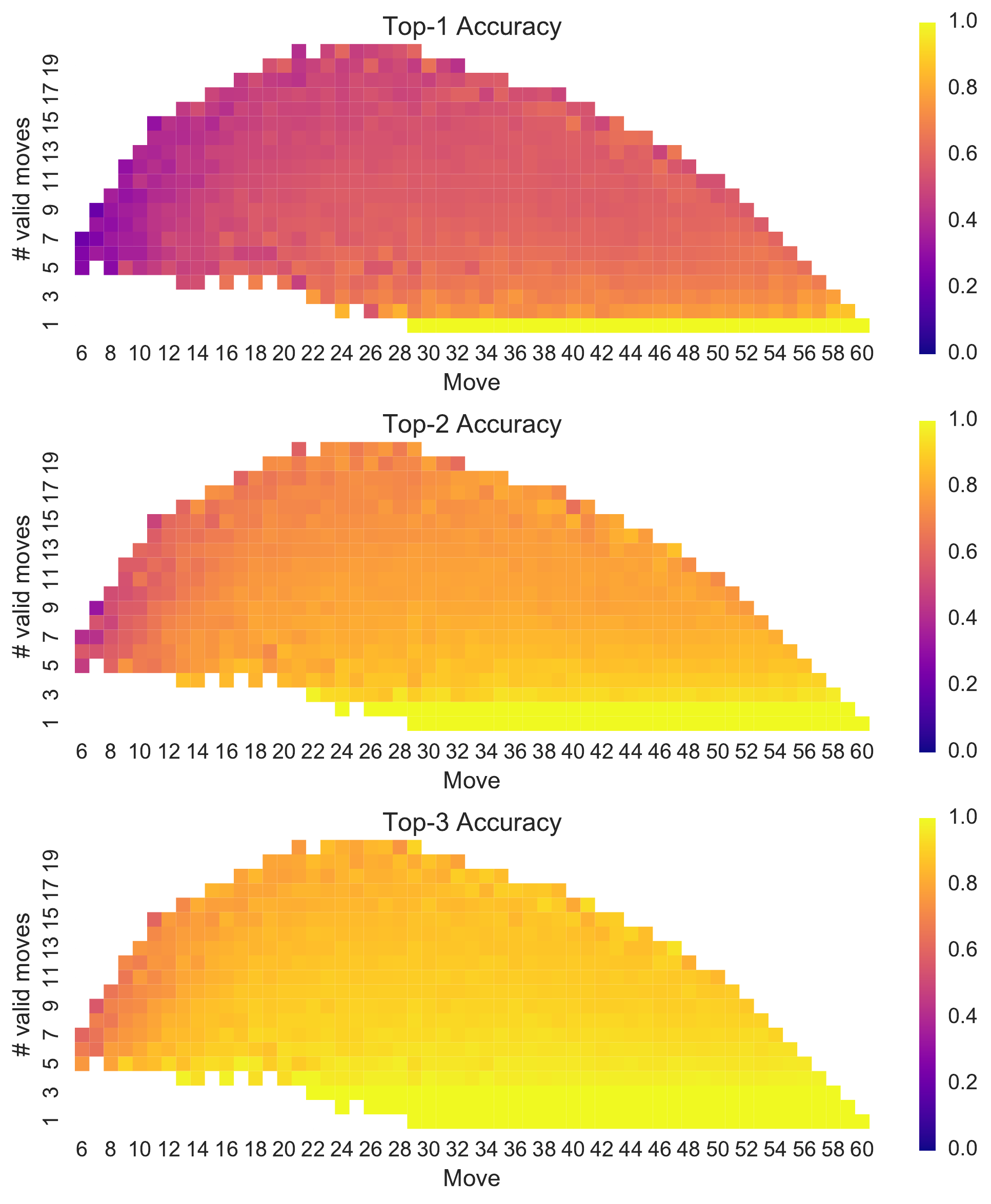}\caption{Top-$k$ prediction accuracy factored w.r.t. the number of move in
a game and the number of valid moves. \label{fig:Top-k-factored}}
\vspace{-5mm}
\end{figure}
The move prediction task is inherently multimodal: to play well,
correct predictions have to be made along the entire course of the
game, for the board states typical for the initial game stages as
well as those occurring in the endgame. It is thus interesting to
ask how difficult it is for a network to perform well at particular
stages of the game. Here, we answer this question in terms of prediction
accuracy. In Fig. \ref{fig:Top-k-factored}, we present the top-$k$
prediction accuracy for our baseline network Conv8+BN, factored with
respect to the number of a move in a game and the number of legal
moves in a given state. Given four pieces in the initial board state,
the former factor may vary from $5$ to $60$; however we start with
$6$ as the fifth move is always the same under the Othello symmetries
(which we take into account here). The color of a datapoint reflects
the probability that the correct move (as per the WThor test set)
is among the top $k$ indicated by the network. 

The figure reveals that the prediction accuracy remains largely consistent
throughout the game. Only in the game's beginning the network's predictions
are less accurate, which we attribute to a relatively small amount
of training data for \emph{Unique-S} at this game stage (due to many
identical states). The other aspect at play, suggested by the differences
between the top-$1$ and top-$3$ graphs, is the large fraction of
inconsistent examples for the early game states, when there are many
alternative paths to success and players tend to follow them accordingly
to their personal habits and preferences. The high quality of predictions
in the later game stages (also when the number of available moves
is high) is particularly encouraging, as even a single move towards
the end of a game can drastically change the state of the board \textendash{}
the feature that Othello owes its name to.

\section{Playing Strength Analysis\label{sec:ExpPlayingStrength}}

Prediction accuracy considered in the previous section is only a surrogate
measure of performance, since we are ultimately interested in the
playing strength. In this section, we let our predictors play against
other players.

\subsection{Experimental Setup}

We consider a suite of thirteen $1$-ply players proposed in previous
research, gathered and compared in our former study \cite{Jaskowski2016cocmaes}.
The suite consists of players with board evaluation functions encoded
by weighted piece counter and n-tuple networks, trained by different
methods including hand-design, temporal difference learning, evolution,
and coevolution (cf. Section \ref{sec:Related-Work}). In addition
to that suite of opponents, we consider also Edax ver. 4.3.2\footnote{http://abulmo.perso.neuf.fr/edax/4.3/edax.4-3-2.zip},
a strong, open source, highly optimized Othello program. Edax can
be configured to play at a given ply (search depth) $n$, which we
denote as Edax-$n$.

Since our move predictors as well as all $14$ opponents (including
Edax) are deterministic, we resort to the following technique in order
to obtain robust estimates of the odds of winning. Rather than playing
a single game that starts from the initial board, we use a set of
$1000$ $6$-ply opening positions prepared by Runarsson and Lucas
\cite{Runarsson2014Preference}. Starting from each such state, two
games are played, with the players switching roles. For a single game,
a player can obtain $1$, $0.5$ or $0$ points on the win, draw,
or loss, respectively. We report the \emph{winning rate}, i.e. the
percentage of points possible to score in $2000$ games.

\subsection{Playing Strength Evaluation\label{subsec:PlayingStrength}}

We selected a representative sample of the five best predictors from
Section \ref{sec:ExpMovePred} and confronted them with all $13$
opponents from our suite as well as with Edax-$1$. Table \ref{tab:round-robin}
presents their average winning rates against the 13 players, and separately
the winning rates against CoCMAES-4+2x2 that proved the best in \cite{Jaskowski2016cocmaes}
and against Edax-$1$.

All networks turn out to be significantly stronger than any of the
existing $1$-ply look-ahead players, achieving around 90\% winning
rate against them. We find this interesting, given that move predictors
are essentially $0$-ply strategies, as they do not perform any look-ahead.
Conv8+BN achieves the highest average winning rate, and is only slightly
worse than Conv8+BN+bagging when playing against CoCMAES-4+2x2 and
than WideNet when playing against Edax-$1$. This overall performance
corroborates its high prediction accuracy. 

\begin{figure}
\selectlanguage{english}%
\begin{centering}
\includegraphics[bb=20bp 510bp 565bp 831bp,clip,width=1\columnwidth]{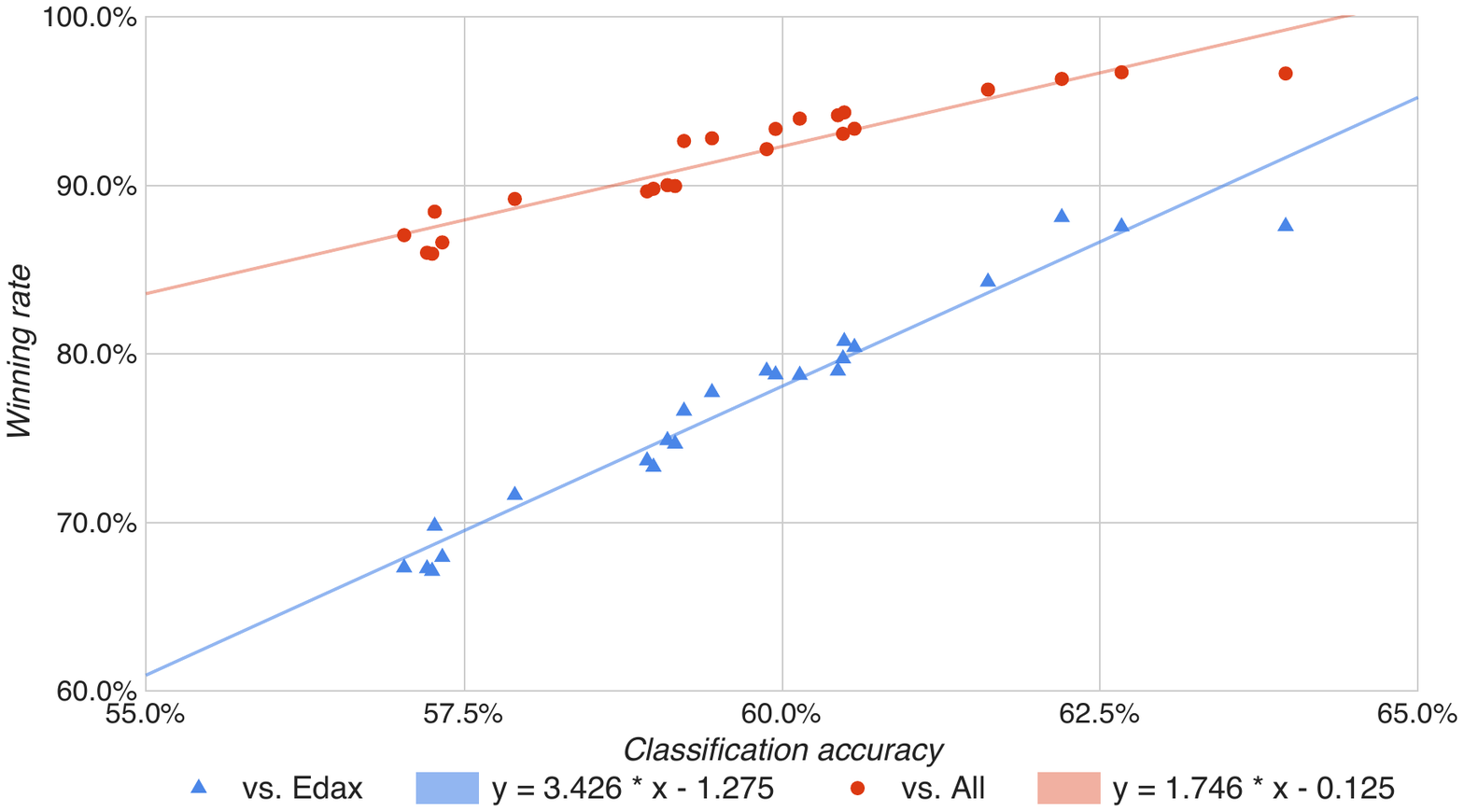}
\par\end{centering}
\caption{\label{fig:acc-relation}Networks' move prediction accuracy vs. winning
rate against all 13 opponent players (red dots) and against Edax (blue
triangles).}

\centering{}\vspace{-5mm}\selectlanguage{american}%
\end{figure}

The roughly monotonous relationship between the prediction accuracy
and the winning rate observed in Table \ref{tab:round-robin} lets
us ask whether such a tendency holds in general. In Fig.~\ref{fig:acc-relation},
we plot the latter against the former for all networks evaluated in
this paper, when confronted with the models in the suite and with
Edax-$1$. The graph reveals a very strong linear dependency (determination
coefficient $r^{2}=0.997$ and $r^{2}=0.999$, respectively). Fitting
linear models suggests that in the considered range of prediction
accuracy ($57$-$64$ percent), each percent point of the prediction
accuracy brings $1.7$ percent points of the winning rate against
the suite of $13$ players, and $3.4$ percent points against Edax-$1$.
Obviously, this trend cannot hold globally, as illustrated by Conv8+BN+bagging
(the rightmost two data points). Although this might be an outlier,
it may also suggest that $63$ percent is approximately the point
at which further increases of prediction accuracy do not translate
into better winning rates anymore. 

\begin{table}
\caption{\label{tab:round-robin}The winning rates of the best performing move
predictors against the $1$-ply opponent players. Note that while
CoCMAES is included in the $13$ opponents suite, Edax-1 is not.}

\centering{}\setlength{\tabcolsep}{3.5pt}%
\begin{tabular}{lcccc}
\hline 
\multirow{3}{*}{Architecture} & Prediction  & \multicolumn{3}{c}{Winning rate}\tabularnewline
\cline{3-5} 
 & accuracy & Average & CoCMAES & \multirow{2}{*}{Edax-$1$}\tabularnewline
 &  & (13 opponents) & -4+2x2 & \tabularnewline
\hline 
ResNet-56-np-p & 60.5 & 94.3 & 87.7 & 80.7 \tabularnewline
DenseNet-sym-100 & 61.9 & 94.1 & 87.5 & 79.0\tabularnewline
WideNet & 62.2 & 96.3 & 92.2 & \textbf{88.1}\tabularnewline
Conv8+BN & 62.7 & \textbf{96.7} & 93.2 & 87.5\tabularnewline
Conv8+BN+bagging & \textbf{64.0} & 96.6 & \textbf{93.5} & 87.6\tabularnewline
\hline 
\end{tabular}
\end{table}

\selectlanguage{american}%

\selectlanguage{english}%

\subsection{Effect of Data Augmentation on Playing Strength}

\begin{table}
\caption{\label{tab:convVsSuite}The winning rates against the 13 players from
the suite for the basic neural predictors trained on different datasets }

\begin{centering}
\setlength{\tabcolsep}{3.5pt}
\par\end{centering}
\centering{}%
\begin{tabular}{lcccc}
\hline 
Architecture & \emph{Original} & \emph{Original-S} & \emph{Unique} & \emph{Unique-S}\tabularnewline
\hline 
Conv4 & 77.9 & 85.2 & 80.5 & \textbf{86.0}\tabularnewline
Conv6 & 84.6 & 89.0 & 86.5 & \textbf{90.0}\tabularnewline
Conv8 & 85.4 & 91.8 & 88.6 & \textbf{93.1}\tabularnewline
\hline 
\end{tabular}
\end{table}

In Section \ref{subsec:PredictionDataAug}, we found out that the
training set stripped of duplicates and augmented with symmetries
(\emph{Unique-S}) gives rise to the best move prediction accuracy.
However, prediction accuracy is only a proxy of what we really care
about \textendash{} the winning rate. It is thus not obvious whether
an analogous claim holds for playing strength, so, in this section
we let the basic convolutional predictors from Section \ref{subsec:MovePredResults}
play against the opponents from the suite. Other settings remain as
in the previous section. 

The winning rates presented in Table \ref{tab:convVsSuite} corroborate
the outcomes in terms of prediction accuracy presented earlier in
Table \ref{tab:augment}. Both duplicate removal and augmentations
have significantly positive effects on the playing strength. This
correlation shows again that, at least in certain intervals, move
prediction accuracy is a reliable predictor of the playing strength
(cf. Fig. \ref{fig:acc-relation}).

\subsection{Playing against Game Tree Search\label{subsec:Edax-n}}

\begin{table}
\caption{\label{tab:edax-search-depth}The performance of Conv8+BN vs. Edax-$n$.}

\begin{centering}
\setlength{\tabcolsep}{3.5pt}
\par\end{centering}
\centering{}%
\begin{tabular}{cccc}
\hline 
 $n$ &  Winning rate & Edax move time {[}$\times10^{-4}$s{]} & Edax prediction accuracy\tabularnewline
\hline 
1 & 87.5 & 1.0 & 47.9\tabularnewline
2 & 59.4 & 1.1 & 53.3\tabularnewline
3 & 34.7 & 1.6 & 55.2\tabularnewline
4 & 19.8 & 5.0 & 59.8\tabularnewline
5 & 10.3 & 10.0 & 59.7\tabularnewline
\hline 
\end{tabular}
\end{table}

Section \ref{subsec:PlayingStrength} revealed that the $0$-ply neural
move predictors outperform the $1$-ply opponents, and do so by a
large margin ($\sim90$ percent vs. the $50$ percent tie point).
It is interesting to see how that margin decreases when the opponents
are allowed for deeper game-tree searches. To investigate this, we
confront Conv8+BN with Edax-$n$ for search depth $n\in[2,5]$ (using
the same protocol involving $2000$ games), and gather the results
in Table \ref{tab:edax-search-depth}. Remarkably, our predictor maintains
superiority when Edax is allowed for $2$-ply lookahead, though the
odds of its winning are now only roughly $6$:$4$. Starting from
search depth $3$, Conv8+BN is more likely to lose than to win, and
for $5$-ply search its odds for winning drop to $1$:$9$. This is
however still impressive, given that the average branching factor
of Othello is $10$, so for search depth $5$ Edax's decision are
based of tens of thousands of board states each, while our neural
predictor performs quite well with just one. 

Search depth impacts the mean time required by Edax to make a move,
which we report in the last column of the table. As Edax is highly
optimized, it is able to make moves at a rate comparable to the prediction
time of the Conv8 CNN (1.9$\times10^{-4}$s) for search depths up
to $3$. Deeper search takes Edax longer. On the other hand, however,
the times reported for neural predictors have been assessed in \emph{batches}
of 256 board states, the common efficiency means in GPU computing.
When asked to predict a move for a single board state, CNNs are 5-20
times slower.

Table \ref{tab:edax-search-depth} includes also Edax prediction accuracy
on the \emph{Original} dataset, which clearly grows with the depth
of search. However, increasing the depth from $4$ to $5$ does not
improve the accuracy, despite the associated increase in head-to-head
matches against Conv8+BN. Also, Edax's prediction accuracy ($59$\%
at $n=5$) is significantly lower than the $64$\% demonstrated by
Conv8+BN. It seems thus that Edax plays in a different way than human
players, whom CNNs try to mimic.

\subsection{Analysis of Game Stages\label{subsec:Hybrid-approach}}

As much as we find the performance of our $0$-ply move predictors
impressive, the previous experiments show that deeper lookaheads are
essential for achieving high winning rates. In our last experiment,
we seek to find out which game stage may benefit the most when supplementing
our move predictor with game tree search (and conversely, which game
stage is most problematic for neural predictors). To this aim, we
split the game into four stages according to intervals of move numbers:
$[1,15]$, $[16,30]$, $[31,45]$, and $[46,60]$. For each stage,
we design a hybrid strategy that employs Edax-$5$ in that stage,
while using the Conv8+BN policy to make decisions in the remaining
stages. We let those hybrid strategies play against Edax-$1$..$5$. 

\begin{figure}
\selectlanguage{english}%
\begin{centering}
\includegraphics[width=1\columnwidth]{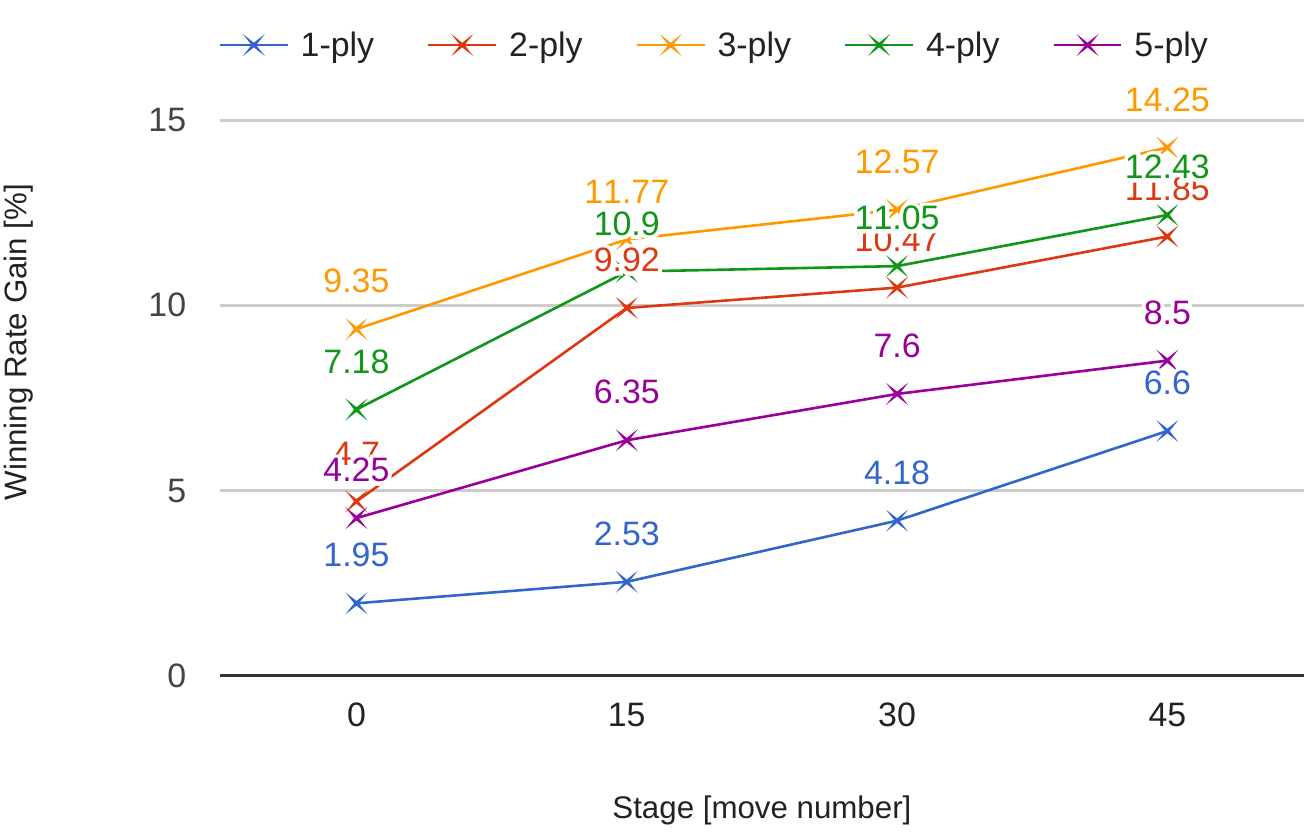}
\par\end{centering}
\caption{\label{fig:gain}The gain of winning rate of Conv8+BN when hybridized
with Edax-5 for a particular stage of the game. Winning rates are
assessed independently against Edax-$1$..$5$. }
\selectlanguage{american}%
\end{figure}

The results summarized in Fig. \ref{fig:gain} confirm that letting
Edax make moves at any game stage improves the performance of the
hybrid. The gains are largest for $n=3$; as demonstrated in Section
\ref{subsec:Edax-n}, for smaller $n$ Edax is relatively easy to
beat already by pure Conv8+BN, so extending it with Edax-$5$ at a
single game stage does not improve the odds for winning much. Conversely,
for $n>3$ the opponent Edax-$n$ becomes so strong that it is hard
to increase the odds of winning either. 

The graphs reveal also that there is more to gain at later stages
of the game regardless of the opponent's depth. We suspect that this
is due to the combinatorial nature of the game, which makes the CNN's
pattern recognition capabilities less relevant for success in the
last game stage. Also, given that Othello games last \emph{at most}
$60$ moves, a search depth $5$ in the last game stage enables Edax
to often reach the final game states and so obtain robust assessments
of potential moves. Last but not least, certain fraction of moves
derived from WThor can be suboptimal and so might have `deceived'
the networks. 

\selectlanguage{american}%

\selectlanguage{english}%

\section{Discussion\label{sec:Discussion}}

The deep CNN-based move predictor with its nearly $63\%$ of accuracy
turned to be much better than n-tuples-based preference learning \cite{Runarsson2014Preference},
which achieved only $53\%$ on the same dataset. Our Conv-8-BN wins
against it in $93.9\%$ of games. We obtained a similar winning rate
of $93.2\%$ also against the CoCMAES player \cite{Jaskowski2016cocmaes},
the strongest to-date $1$-ply Othello player (cf. Table \ref{tab:round-robin}). 

Despite Othello's relatively low branching factor and small board,
CNNs proved to be supreme move predictors, managing to efficiently
capture the expertise materialized in millions of humans' moves. We
find that impressive given that small, discrete-valued board states
are arguably very different from natural images\footnote{Note that the images considered in computer vision hardly ever contain
fewer than 100 pixels.}. 

Seeing a \emph{convolutional} network to perform so well in a board
game that is characterized with little translation invariance is intriguing,
as the main motivation for using convolution is precisely the translation
invariance. This suggests that certain patterns observed in Othello
boards can be to some extent translation-invariant, or at least can
be detected using `higher-order convolutions' implemented by a stack
of several convolutional layers intertwined with nonlinearities.

Concerning limitations of the approach, we could not improve upon
the result obtained with the Conv-8-BN network by using deeper networks
with skip connections. This suggests that what works for high-dimensional
image data does not always succeeds for low-dimensional problems such
as Othello. 

An interesting observation concerning training is that the large
networks hardly overfit to the data. The classification accuracy on
the training set is typically within one percent point of the accuracy
on the test set, suggesting that the networks we considered here cannot
memorize the training data entirely. We may conclude thus that there
is still room for improving their architectures. In particular, this
may be a sign of an inherent limitation of convolutional processing,
where the limited in size RFs may be insufficient to capture arbitrary
complex combinatorial patterns occurring in entire board states. A
natural follow-up step is thus considering fully-connected networks
that do not suffer from that limitation; however, they may be also
much more costly in training and tend to overfit. In relation to that,
an interesting study regarding fully-connected networks for Othello
with up to 3 hidden layers was performed in \cite{Binkley2007Study}.

\section{Conclusions}

Our study brings evidence that deep neural networks are a viable methodology
for training Othello agents. They do not require explicit knowledge
on game rules or strategies, nor extensive game-tree searches, nor
explicitly compare individual future board states or actions. Rather
than that, they achieve a high level of playing by learning the associations
between the patterns observed in board states and the desired moves.
We may claim thus that, at least to some extent, the trained neural
predictors embody the knowledge of the Othello domain \textendash{}
a statement that would be hard to defend for game-tree search techniques,
which owe their power largely to algorithmic efficiency. 

On the other hand, the supervised approach proposed here cannot be
deemed as `knowledge-free' in the same sense as for instance evolutionary
or reinforcement learning methods. Here, the essential know-how is
hidden in the expert games used for training. CNNs power lies in the
ability to extract and incorporate this knowledge into a system that
is able to generalize beyond the training examples, something no other
method in the past has been shown to do well enough to achieve the
expert level of play in Othello.

Given the supreme performance of CNNs, this work may be worth extending
in several directions. The networks that rely more on the fully-connected
layers, mentioned in Section \ref{sec:Discussion}, are one such possibility.
Logistello \cite{Buro1997OthelloMultiProbCut} and Edax use different
evaluation functions for each stage of the game (cf. Section\ref{subsec:Hybrid-approach}),
and such `factorization by game course' could also lead to a better
CNN agent. Another, arguably simple extension is to add the information
about the player to move, which we expect to be beneficial given that
Othello is a non-symmetric game. Last but not least, given the usefulness
of convolutional features evidenced here, it would be interesting
to verify their capabilities in other contexts than supervised learning
of move predictors, e.g. as state or state-action evaluation functions
trained with reinforcement learning methods. 

\section*{Acknowledgments}

P. Liskowski acknowledges the support from grant 2014/15/N/ST6/04572
funded by the National Science Centre, Poland. W. Ja\'{s}kowski acknowledges
the support from Ministry of Science and Higher Education grant ``Mobility
Plus'' no 1296/MOB/IV/2015/0. K.\,Krawiec was supported by the National
Science Centre, Poland, grant no.~2014/15/B/ST6/05205.

\bibliographystyle{plain}
\bibliography{all,wjaskowski,library}

\end{document}